\IfFileExists{ieeeconf.cls}{%
  \documentclass[letterpaper,10pt,conference]{ieeeconf}
  \IEEEoverridecommandlockouts
  \overrideIEEEmargins
  \addtolength{\topmargin}{10pt}
}{%
  \documentclass[conference,letterpaper]{IEEEtran}
  \IEEEoverridecommandlockouts
}

\usepackage[T1]{fontenc}
\usepackage[utf8]{inputenc}
\usepackage{microtype}
\usepackage{etoolbox}
\usepackage{graphicx}
\usepackage{subcaption}
\usepackage{amsmath,amssymb,amsfonts,bm}
\usepackage{booktabs}
\usepackage[table]{xcolor}
\usepackage{multirow}
\let\labelindent\relax
\usepackage{enumitem}
\usepackage{listings}
\usepackage[hidelinks]{hyperref}
\usepackage[nameinlink,capitalize]{cleveref}
\usepackage{mathtools}
\usepackage{placeins}
\usepackage{array}

\crefname{figure}{Fig.}{Figs.}
\Crefname{figure}{Fig.}{Figs.}
\crefname{subfigure}{Fig.}{Figs.}
\Crefname{subfigure}{Fig.}{Figs.}

\makeatletter
\@ifundefined{keywords}{%
}{}
\makeatother


\makeatletter
\begingroup
\lccode`\!=`\-
\lowercase{\endgroup
  \patchcmd{\abstract}{!!!}{:\ }{}{}
  \patchcmd{\IEEEkeywords}{!!!}{:\ }{}{}
}
\makeatother

\definecolor{SMILERow}{RGB}{226,242,255}
\makeatletter
\def\@IEEEauthorblockAtopspace{0.45em}
\makeatother

\title{\LARGE \bf
SMILE: Smooth Motion for Improved Long-Horizon VLA Execution
}
\author{%
\authorblockN{Jongwoo Park\textsuperscript{1}, E-Ro Nguyen\textsuperscript{1}, Kanchana Ranasinghe\textsuperscript{2},\\
Cristina Mata\textsuperscript{1}, Xiang Li\textsuperscript{1}, and Michael S Ryoo\textsuperscript{1}}
\authorblockA{\textsuperscript{1}Stony Brook University \quad \textsuperscript{2}Salesforce AI Research\\[0.45em]
{\tt\small jongwopark@cs.stonybrook.edu}}
\thanks{This work has been submitted to the IEEE for possible publication. Copyright may be transferred without notice, after which this version may no longer be accessible.}
}

\begin{document}

\maketitle
\thispagestyle{empty}
\pagestyle{empty}
\raggedbottom

\begin{abstract}
Vision-Language-Action (VLA) models reduce inference cost by executing multiple actions per call, but longer horizons often degrade accuracy because raw chunks contain jitter and outliers. We introduce SMILE, an architecture-preserving interface that predicts B-spline coefficients and decodes them into smooth action sequences. SMILE changes only the action representation, enabling longer fixed horizons while retaining each baseline's backbone and model scale. We apply SMILE to SmolVLA, Evo1, VPP, and DAWN, improving accuracy and amortized inference efficiency across LIBERO, CALVIN, and real-world experiments. SMILE-Evo1 reaches 98.0\% with a 1.1x speedup on LIBERO, while SMILE-VPP reaches an average length of 4.42 with a 1.5x speedup on CALVIN. At a matched execution horizon of 10, SMILE-SmolVLA reduces non-boundary acceleration by 78.6\% and velocity sign-change rate by 42.3\%. Real-world xArm tests show higher success, fewer drops, and fewer contacts. These results establish smooth coefficient-space generation as a route to accurate, efficient long-horizon VLA execution.
\end{abstract}

\noindent\textbf{Project page:}~{\footnotesize\href{https://jongwoopark7978.github.io/smilevla/}{\texttt{jongwoopark7978.github.io/smilevla}}}

\begin{figure*}[!t]
    \centering
    \begin{subfigure}[b]{0.33\textwidth}
        \centering
        \includegraphics[width=\linewidth]{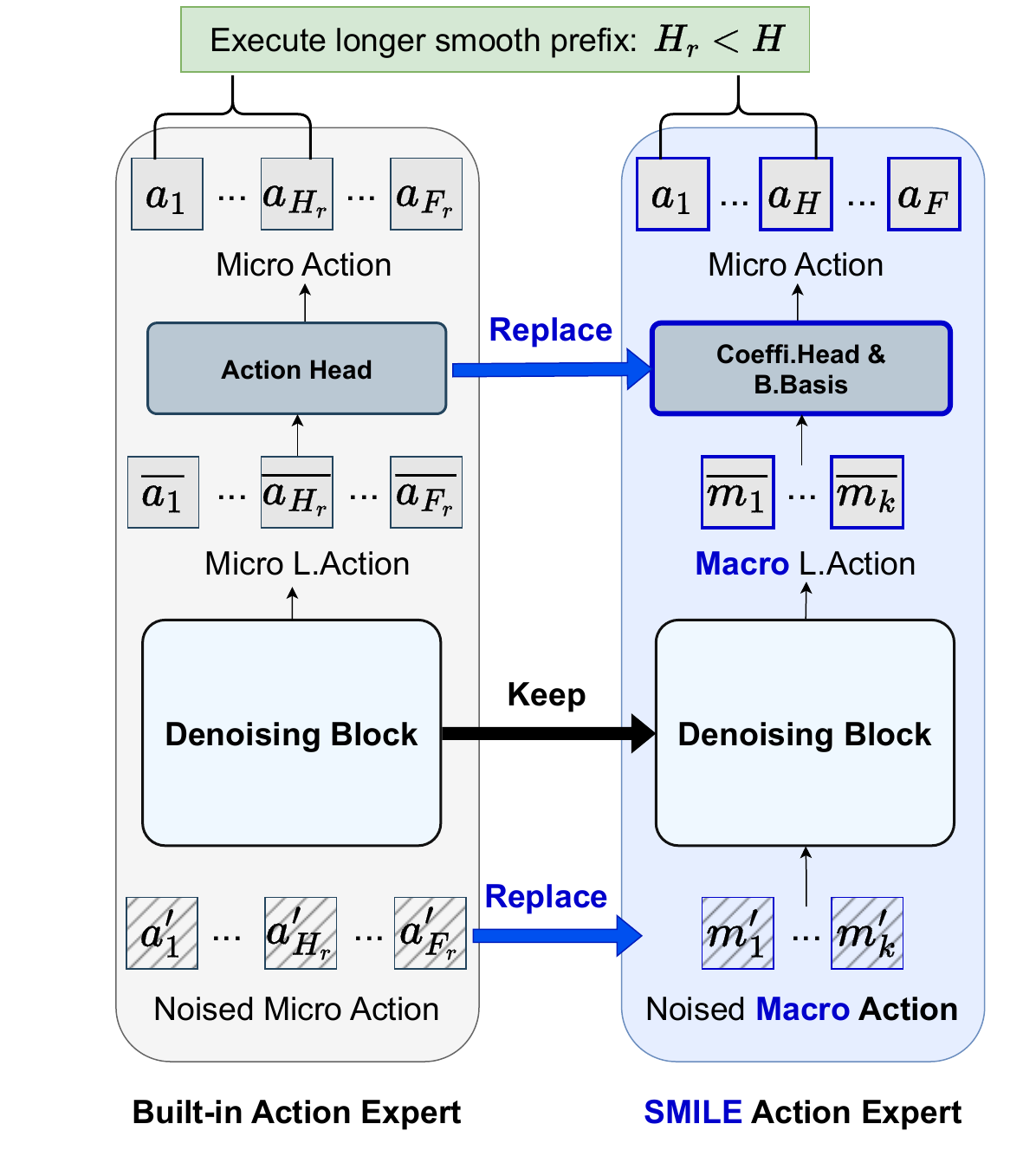}
        \caption{SMILE architecture.}
        \label{fig:smile-architecture}
    \end{subfigure}
    \hfill
    \begin{subfigure}[b]{0.65\textwidth}
        \centering
        \includegraphics[width=\linewidth]{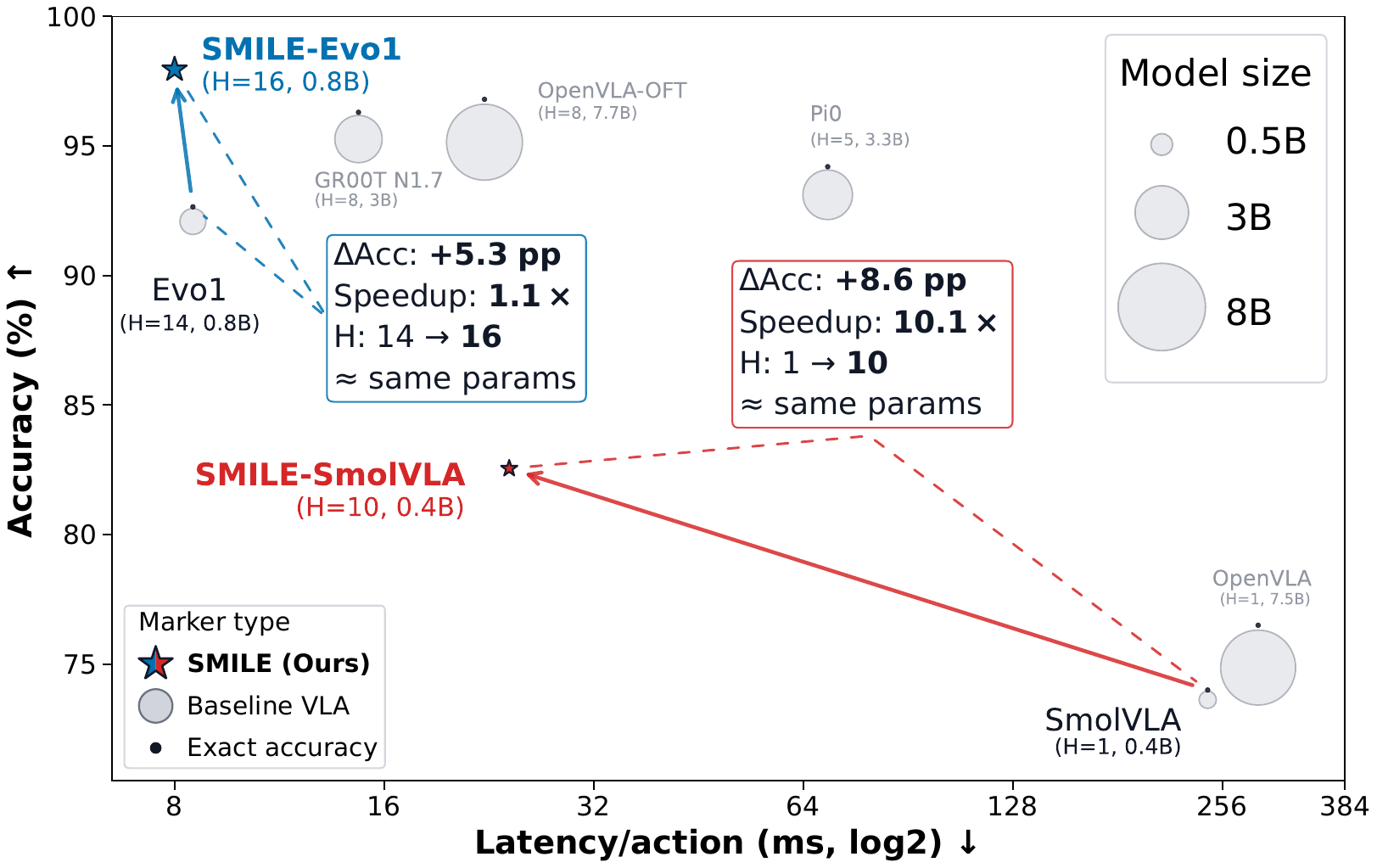}
        \caption{Accuracy-efficiency comparison on LIBERO.}
        \label{fig:pareto-efficiency}
    \end{subfigure}
    \caption{SMILE architecture and accuracy-efficiency frontier. (a) The built-in action expert directly outputs raw micro-actions $a_1,\ldots,a_{F_r}$ and executes a shorter prefix $H_r$, while SMILE replaces only the action-side raw sequence interface with B-spline coefficient tokens and coefficient heads that decode to smooth micro-actions $a_1,\ldots,a_F$, enabling a longer executed prefix $H_r < H$. The original VLA backbone, conditioner, and main denoising action-expert block are retained. (b) Success versus amortized inference latency per action; marker size denotes model scale. SMILE improves the accuracy-efficiency trade-off at comparable model size.}
    \label{fig:architecture-pareto}
\end{figure*}

\section{Introduction}
\label{sec:introduction}
Vision-Language-Action (VLA) models increasingly rely on action chunking to make closed-loop robot control computationally practical. Rather than invoke a large multimodal policy at every controller step, an action expert predicts a short sequence of low-level controls and the robot executes several actions before refreshing the observation \cite{act2023,diffusionpolicy2023,smolvla2025,vpp2024}. The number of actions executed per refresh, which we denote by the execution horizon $H$, directly controls inference efficiency: larger $H$ amortizes one expensive model call over more robot actions.

The benefit of a larger horizon is often offset by a loss in control quality. Raw action chunks can contain high-frequency oscillations, temporally inconsistent directions, and isolated outliers. These errors are tolerable when only one or two predictions are executed, but they accumulate when a longer prefix is executed before replanning. As a result, increasing $H$ can lower amortized latency while simultaneously reducing task success. This behavior is visible for both SmolVLA and Evo1 in \cref{fig:baseline-horizon-motivation}: their official raw-action policies become less accurate as more actions are executed per model call. The central question of this work is therefore how to make the generated action sequence reliable enough that a VLA can safely use a longer fixed horizon.

We propose \emph{SMILE}, a smooth hierarchical action interface for denoising VLA action experts. Instead of denoising every low-level action independently in raw action space, SMILE predicts a compact set of B-spline coefficients for translation, rotation, and gripper control. A fixed basis decodes these coefficients into a temporally coherent sequence of execution actions. The coefficient sequence acts as a macro-action representation, while the decoded controls remain compatible with the original robot controller. Because the smoothness is imposed by the representation rather than by a larger backbone or a separate planner, SMILE can increase the execution horizon and reduce amortized latency without increasing the reported model scale.

SMILE is designed as an architecture-preserving replacement for the action-side interface. We retain the original visual-language conditioner and the main denoising backbone, and replace only the lightweight projections and output heads that map between the denoiser and robot actions. This design applies across heterogeneous VLA action experts: VLM-centered SmolVLA and Evo1 \cite{smolvla2025,evo12026}; predictive-visual VPP \cite{vpp2024}; and explicit pixel-motion-conditioned DAWN \cite{dawn2026}. The upstream predictive visual or pixel-motion modules remain unchanged.

Our experiments show consistent accuracy--efficiency gains across LIBERO, CALVIN, and real-world xArm manipulation. SMILE improves performance across four heterogeneous action experts while enabling longer fixed execution horizons at comparable model-size scales. A matched-horizon analysis further shows substantially lower intra-chunk acceleration and oscillatory action changes, linking these gains to smoother action generation.

\noindent\textbf{Contributions.} We make three main contributions:
\begin{enumerate}[leftmargin=*,nosep]
    \item We introduce SMILE, a B-spline coefficient action interface that generates smooth action sequences and enables VLAs to execute longer fixed horizons with higher task accuracy and lower amortized inference latency per action. This capability is particularly beneficial for temporally extended manipulation tasks, such as those evaluated in LIBERO and CALVIN.
    
    \item We provide an architecture-preserving integration that retains each baseline's original conditioning pathways and denoising backbone, while maintaining a model size comparable to that of the corresponding baseline. This interface supports heterogeneous policy designs, including VLM-centered, predictive-visual, and explicit pixel-motion-conditioned action experts.
    
    \item We demonstrate consistent gains across SmolVLA, Evo1, VPP, and DAWN on LIBERO, CALVIN, and real-world xArm manipulation, and directly verify that these improvements correlate with substantially smoother fixed-horizon action execution.
\end{enumerate}

\section{Related Work}
\label{sec:related-work}

\paragraph{Action chunking and efficient VLA execution}
Action chunking predicts a sequence of future controls and executes a prefix before replanning, amortizing policy inference while trading closed-loop reactivity for fewer model refreshes \cite{act2023,diffusionpolicy2023,smolvla2025,vpp2024}. BID samples and selects candidate chunks at inference time, whereas AAC and AutoHorizon adapt how much of each predicted chunk is executed before replanning \cite{bid2025,aac2026,autohorizon2026}. These methods modify chunk selection or execution without changing the learned action representation. SMILE instead reparameterizes the action-side denoising target into compact B-spline coefficients, improving the temporal quality of each generated chunk so that a longer fixed horizon $H$ can be executed reliably. SMILE therefore targets the accuracy--amortized-inference frontier for long fixed-horizon execution, while remaining complementary to candidate sampling and adaptive-horizon selection.

\paragraph{Smooth action representations and real-time execution}
Action-chunking policies such as ACT and Diffusion Policy predict multi-step controls to improve temporal coherence, although raw action chunks can still exhibit intra-chunk jitter and discontinuities across successive policy updates \cite{act2023,diffusionpolicy2023}. RTC and ABPolicy primarily address real-time asynchronous execution by generating the next chunk while the current chunk is being executed; RTC conditions the new chunk on previously committed actions, whereas ABPolicy combines a B-spline control-point representation with continuity-aware trajectory updates \cite{rtc2025,abpolicy2026}. A complementary line of work develops compact, structured action representations. FAST compresses action sequences in the frequency domain, BEAST formulates fixed-length B-spline tokens and demonstrates them with a vision-language foundation model and non-VLM policies, and recent spline-based policies directly predict continuous spline trajectories \cite{fast2025,beast2025,bsplinepolicy2026,splinepolicy2026}. Within this landscape, SMILE specifically targets synchronous fixed-horizon execution. It provides an architecture-preserving B-spline coefficient interface for heterogeneous compact VLA action experts, retaining their conditioning pathways and main denoising backbones, and explicitly evaluates whether smoother chunks support a longer fixed execution horizon $H$ with higher task accuracy and lower amortized inference latency per action. Thus, SMILE focuses on improving the accuracy--efficiency operating point through reliable long fixed-horizon chunks, rather than primarily addressing inference delay through asynchronous scheduling or introducing a general-purpose action tokenizer.

\paragraph{Denoising action experts and structured visual conditioning}
A growing class of continuous-action VLAs couples a pretrained vision-language or multimodal backbone with a diffusion- or flow-matching action expert that generates complete action chunks \cite{diffusionpolicy2023,smolvla2025,evo12026,pi02025,dexvla2025}. Beyond direct visual-language conditioning, recent policies incorporate dynamics-aware visual structure through predictive video features, compact world-knowledge forecasts, explicit motion intermediates, or joint video-action generation \cite{vpp2024,dawn2026,dreamvla2025,uwm2025, videovla2025,udvla2026}. In our evaluation, VPP conditions an inverse-dynamics policy on predictive representations learned by a video diffusion model, whereas DAWN first generates dense pixel motion and then conditions a low-level diffusion action expert on this explicit motion representation \cite{vpp2024,dawn2026}. Together with the VLM-centered SmolVLA and Evo1, these substantially different designs provide heterogeneous testbeds for assessing whether coefficient-space denoising is tied to a particular architecture. SMILE preserves each policy's original visual-language and motion-conditioning pathways and main denoising backbone, while reparameterizing only the denoised action variable and action-side interface from per-step controls to compact B-spline coefficients. SMILE therefore serves as an architecture-preserving action interface rather than replacing the underlying perceptual or predictive model.

\section{Observations on Long-Horizon Execution}
\label{sec:long-horizon-observation}

\begin{figure}[!t]
    \centering

    \begin{subfigure}[t]{0.60\columnwidth}
        \vspace{-4.5pt}
        \centering
        \includegraphics[width=\linewidth]
        {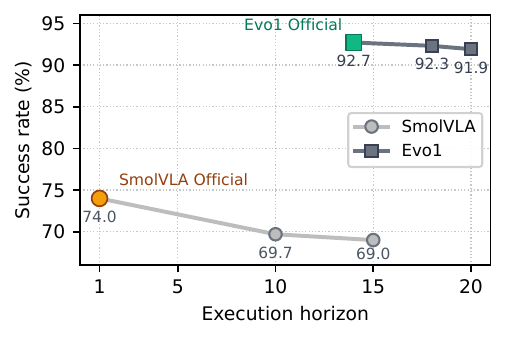}
        \caption{Success vs execution horizon}
        \label{fig:raw-action-horizon-motivation}
    \end{subfigure}
    \hfill
    \begin{subfigure}[t]{0.34\columnwidth}
        \vspace{0pt}
        \centering
        \scriptsize
        \setlength{\tabcolsep}{2.2pt}
        \renewcommand{\arraystretch}{1.08}
        \begin{tabular}{@{}lcc@{}}
            \toprule
            & Base & SMILE \\
            \midrule
            \multicolumn{3}{c}{\textit{Evo1 pair}} \\
            $H$       & 6    & \textbf{16} \\
            DR (\%)   & 8.3  & \textbf{0.0} \\
            HR (\%)   & 45.0 & \textbf{22.5} \\
            \midrule
            \multicolumn{3}{c}{\textit{VPP pair}} \\
            $H$       & 10   & \textbf{15} \\
            DR (\%)   & 3.3  & \textbf{0.0} \\
            HR (\%)   & 25.0 & \textbf{11.7} \\
            \bottomrule
        \end{tabular}
        \caption{Cluttered results.}
        \label{fig:realworld-horizon-motivation}
    \end{subfigure}

    \caption{Motivation for smooth long-horizon execution. 
    (a) LIBERO success of raw-action baselines across execution horizons; colored markers denote their official default horizons.
    (b) Selected cluttered xArm results for each baseline--SMILE pair. $H$ denotes the execution horizon while DR and HR denote drop and hit rates; complete results appear in \cref{tab:realworld-xarm}.}
    \label{fig:baseline-horizon-motivation}
\end{figure}

We first examine whether simply increasing the execution horizon $H$, defined as the number of actions executed per model refresh, is sufficient for efficient long-horizon execution. Although a larger $H$ amortizes each model call over more controller steps, it also increases the number of actions executed before the next observation is processed.

\Cref{fig:baseline-horizon-motivation} presents two complementary
observations. First, the raw-action baselines lose task accuracy as their execution horizons increase, showing that longer execution cannot be obtained reliably by increasing $H$ alone. Second, the SMILE variants operate with longer fixed horizons in the cluttered xArm setting while reducing object drops and contacts with neighboring objects. The lower drop and hit rates are consistent with more stable gripper behavior and more controlled approach and transport motions.

Together, these results motivate improving the temporal structure of the predicted action chunk rather than merely increasing its execution horizon. SMILE addresses this limitation by denoising compact B-spline coefficients whose decoded actions are coupled across time.

\section{Method}
\label{sec:method}
\subsection{Overview}
At model refresh $j$, a standard action expert predicts a raw action chunk
\begin{equation}
A_j=[a_{j,1},\ldots,a_{j,F}]\in\mathbb{R}^{F\times d},
\end{equation}
where $F$ is the generated chunk length. At inference time, the controller executes only the first $H\leq F$ actions before the next observation is processed \cite{smolvla2025,vpp2024,evo12026,dreamvla2025, openvla2024,openvlaoft2025,gr00tn17_2026, seer2024,vidman2024}. Increasing $H$ reduces amortized inference latency per action, but it also commits the controller to a longer locally open-loop prefix whose individual actions may be temporally inconsistent.

SMILE replaces the raw chunk with a compact macro-action
\begin{equation}
M_j=\bigl(\Theta^{\mathrm{tr}}_j,\Theta^{\mathrm{rot}}_j,\Theta^{\mathrm{grip}}_j\bigr),
\end{equation}
consisting of B-spline coefficients for translation, rotation, and gripper control. A deterministic decoder maps $M_j$ to an $F$-step smooth action sequence that is sent to the original controller, and the controller executes the first $H$ actions. Within each modality, the decoded sequence is jointly determined by a compact set of spline coefficients, coupling neighboring actions across time and reducing sensitivity to isolated predictions. The execution horizon $H$ is fixed within each rollout rather than adapted online.

As shown in \cref{fig:smile-architecture}, SMILE retains the computationally dominant components of the baseline, including its VLA backbone, conditioning pathway, and main denoising backbone. Thus, visual-language, predictive-visual, or pixel-motion features are computed as in the original model. We modify only the action-side input projection so that the denoiser operates on noisy coefficient tokens and replace the raw-action output projection with coefficient heads.

\subsection{B-spline action representation and decoding}
For a generated chunk of length $F$, spline degree $P$, and $K$ control points, let $\Phi_F^{(P,K)}\in\mathbb{R}^{F\times K}$ denote the fixed B-spline basis evaluated at the $F$ uniformly spaced action times. For each chunk start $s$, the action expert predicts
\begin{equation}
\Theta_s^{\mathrm{tr}}\in\mathbb{R}^{K\times3},\qquad
\Theta_s^{\mathrm{rot}}\in\mathbb{R}^{K\times3},\qquad
\Theta_s^{\mathrm{grip}}\in\mathbb{R}^{K\times1}.
\end{equation}
The cumulative translation, cumulative command-space rotation, and continuous gripper trajectories are decoded as
\begin{equation}
\begin{aligned}
P_s &= \Phi_F^{(P,K)}\Theta_s^{\mathrm{tr}},\\
R_s &= \Phi_F^{(P,K)}\Theta_s^{\mathrm{rot}},\\
G_s &= \operatorname{clip}\!\left(
    \Phi_F^{(P,K)}\Theta_s^{\mathrm{grip}},g_{\min},g_{\max}
\right).
\end{aligned}
\end{equation}
Here, $g_{\min}$ and $g_{\max}$ denote the benchmark- or controller-specific lower and upper bounds for the gripper command.

Using the virtual initial cumulative states $P_s(0)=\mathbf{0}$ and $R_s(0)=\mathbf{0}$, translation actions are finite differences of the cumulative path:
\begin{equation}
a^{\mathrm{tr}}_{s,\tau}=P_s(\tau)-P_s(\tau-1).
\end{equation}
For rotation, we map the cumulative rotation vectors to $SO(3)$ and take adjacent relative rotations:
\begin{equation}
a^{\mathrm{rot}}_{s,\tau}
=
\log\!\left(
\exp(R_s(\tau-1))^{-1}\exp(R_s(\tau))
\right),
\end{equation}
which is converted back to the controller's rotation-vector convention. The gripper command is $a^{\mathrm{grip}}_{s,\tau}=G_s(\tau)$. 

The decoded actions $a_{s,\tau}$ are native controller actions; SMILE changes only the representation predicted by the action expert before deterministic decoding. Translation and rotation retain the baseline's relative-action semantics, while the gripper command remains absolute. At inference time, the controller executes the first $H\leq F$ decoded actions. Unless otherwise stated, all SMILE variants use $K=8$ control points and cubic splines ($P=3$), as selected in \cref{tab:spline-ablation}.

\subsection{Architecture-preserving denoising objective}
SMILE changes the variable denoised by the action expert from a raw action sequence to normalized spline coefficients. Let $\Theta_{s,\mathrm{norm}}^{m,\star}$ denote the target coefficient tensor for modality $m\in\{\mathrm{tr},\mathrm{rot},\mathrm{grip}\}$, and let $c_s$ denote the baseline's unchanged multimodal conditioning. For each modality, the baseline's native corruption process can be written generically as
\begin{equation}
\Theta_{s,t}^{m}
=
\alpha_t\Theta_{s,\mathrm{norm}}^{m,\star}
+
\sigma_t\epsilon_m,
\qquad
\epsilon_m\sim\mathcal{N}(0,I).
\end{equation}
The preserved denoiser predicts the target used by the original training formulation: a velocity field for flow-matching experts or noise for diffusion experts. With target $y_t^m$, the objective is
\begin{equation}
L_{\mathrm{denoise}}^m
=
\left\|
f_\theta^m(\Theta_{s,t},t,c_s)-y_t^m
\right\|_F^2,
\end{equation}
\begin{equation}
L_{\mathrm{total}}
=
\lambda_{\mathrm{tr}}L_{\mathrm{denoise}}^{\mathrm{tr}}
+
\lambda_{\mathrm{rot}}L_{\mathrm{denoise}}^{\mathrm{rot}}
+
\lambda_{\mathrm{grip}}L_{\mathrm{denoise}}^{\mathrm{grip}}.
\label{eq:total-loss-main}
\end{equation}
We use equal coefficient-denoising weights for all SMILE variants, setting $\lambda_{\mathrm{tr}}=\lambda_{\mathrm{rot}}=\lambda_{\mathrm{grip}}=1$.

\begin{figure}[!t]
    \centering
    \begin{minipage}[t]{0.34\columnwidth}
        \centering
        \includegraphics[width=\linewidth]{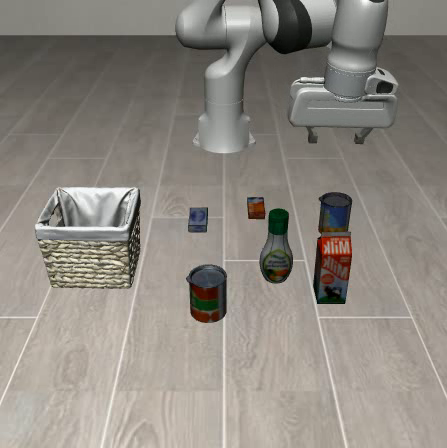}
    \end{minipage}
    \hspace{0.03\columnwidth}
    \begin{minipage}[t]{0.34\columnwidth}
        \centering
        \includegraphics[width=\linewidth]{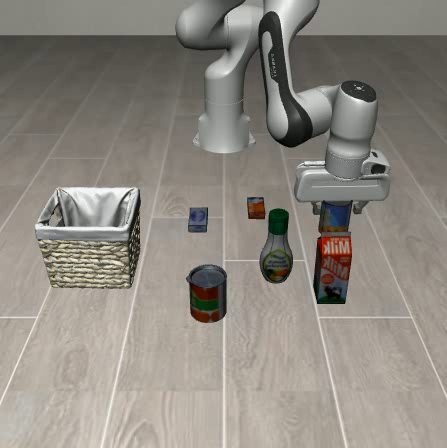}
    \end{minipage}\\[-0.35em]
    \begin{minipage}[t]{0.34\columnwidth}
        \centering
        \includegraphics[width=\linewidth]{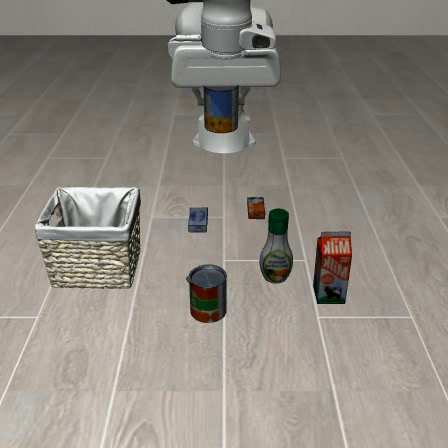}
    \end{minipage}
    \hspace{0.03\columnwidth}
    \begin{minipage}[t]{0.34\columnwidth}
        \centering
        \includegraphics[width=\linewidth]{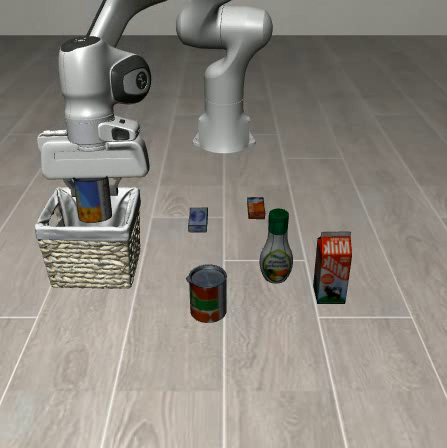}
    \end{minipage}
    \caption{SMILE-Evo1 success on LIBERO-Long: Each image shows approach, grasp, lift, and place.}
    \label{fig:libero-long-success}
\end{figure}

\subsection{Coefficient-target construction}
\label{sec:coefficient-target-construction}

We construct coefficient targets from the same expert action trajectories used by each baseline. For every eligible start step $s$, we collect the next $F$ native actions, convert translation and rotation into cumulative command-space trajectories, and retain the gripper commands in their absolute form. For each modality $m\in\{\mathrm{tr},\mathrm{rot},\mathrm{grip}\}$, the target coefficients are obtained by least-squares projection onto the fixed B-spline basis:
\begin{equation}
\Theta_s^{m,\star}
=
\arg\min_{\Theta}
\left\|
\Phi_F^{(P,K)}\Theta-Y_s^m
\right\|_F^2,
\end{equation}
where $Y_s^m$ denotes the corresponding cumulative or absolute target trajectory. We compute dataset-level coefficient statistics for normalization and apply validity masks near episode boundaries.

In summary, SMILE replaces raw per-step action denoising with coefficient-space denoising. Target coefficients are obtained by least-squares projection of transformed trajectories onto a fixed B-spline basis. At inference time, the denoised coefficients are decoded through the same basis and generate temporally coupled native controller actions.

\section{Experiments}
\label{sec:experiments}

We evaluate SMILE across four architecturally heterogeneous action experts on LIBERO, CALVIN, and real-world xArm environments, measuring task accuracy, execution horizon, amortized inference latency, and action smoothness.

\subsection{LIBERO Benchmark}
\label{sec:libero-experiments}

LIBERO contains four suites, Spatial, Object, Goal, and Long, comprising 40 language-conditioned manipulation tasks \cite{libero2023}. We evaluate SmolVLA and Evo1 over 2,000 episodes, using 50 episodes per task, and report suite-wise and overall success together with execution horizon, amortized inference latency, and model size. LIBERO-Long specifically evaluates temporally extended manipulation, where errors within an executed action chunk can compound across task stages. A representative SMILE-Evo1 success rollout is shown in \cref{fig:libero-long-success}.

\begin{table*}[t]
\centering
\caption{SMILE variants improve accuracy with longer fixed execution horizons at comparable model scale on LIBERO. Inf./action denotes amortized inference latency per executed action.}
\label{tab:libero-horizon-metrics}
\scriptsize
\setlength{\tabcolsep}{3pt}
\resizebox{\textwidth}{!}{%
\begin{tabular}{lcccccccc}
\toprule
Model & Exe. Horizon $\uparrow$ & Spatial $\uparrow$ & Object $\uparrow$ & Goal $\uparrow$ & Long $\uparrow$ & All $\uparrow$ & Inf./action (ms) $\downarrow$ & Model Size $\downarrow$ \\
\midrule
GR00T N1.7 \cite{gr00tn17_2026} & 8 & 97.2 & 96.8 & 97.8 & 93.4 & 96.3 & 14.7 & 3B \\
$\pi_0$ \cite{pi02025} & 5 & 96.8 & 98.8 & 95.8 & 85.2 & 94.2 & 69.4 & 3.3B \\
OpenVLA \cite{openvla2024} & 1 & 84.7 & 88.4 & 79.2 & 53.7 & 76.5 & 288.2 & 7.5B \\
OpenVLA-OFT \cite{openvlaoft2025} & 8 & 97.7 & 98.0 & 96.1 & 96.1 & 96.8 & 22.3 & 7.7B \\
\midrule
SmolVLA \cite{smolvla2025} & 1 & 78.8 & 91.4 & 82.8 & 42.8 & 74.0 & 243.9 & 0.4B \\
\rowcolor{SMILERow}
SMILE-SmolVLA (Ours) & 10 & 84.8 & 95.0 & 86.8 & 63.6 & 82.6 & 24.2 & 0.4B \\
\midrule
Evo1 \cite{evo12026} & 14 & 88.4 & 94.2 & 98.2 & 89.8 & 92.7 & 8.5 & 0.8B \\
\rowcolor{SMILERow}
SMILE-Evo1 (Ours) & 16 & 97.2 & 99.6 & 96.8 & 98.2 & 98.0 & 8.0 & 0.8B \\
\bottomrule
\end{tabular}%
}
\end{table*}

\Cref{fig:pareto-efficiency,tab:libero-horizon-metrics} summarize the LIBERO accuracy--efficiency results. SMILE-Evo1 increases overall success from 92.7\% to 98.0\% while extending the execution horizon from 14 to 16 and reducing amortized latency from 8.5 to 8.0~ms per action. SMILE-Evo1 therefore establishes the strongest accuracy-efficiency operating point among the compared methods. SMILE-SmolVLA increases success from 74.0\% to 82.6\% while extending the horizon from 1 to 10 and reducing latency from 243.9 to 24.2~ms per action, corresponding to a 10.1$\times$ speedup. Both integrations retain the reported model-size scales of their corresponding baselines.

The suite-level results show that SMILE-SmolVLA improves all four suites, with its largest gain on LIBERO-Long, from 42.8\% to 63.6\%. SMILE-Evo1 also improves Spatial, Object, and Long, reaching 98.2\% on LIBERO-Long. These gains support the hypothesis that smoother action generation is particularly beneficial for temporally extended tasks requiring reliable execution over longer fixed horizons.

Unlike the raw-action trade-off in \cref{sec:long-horizon-observation}, these paired comparisons improve accuracy and inference efficiency simultaneously, without increasing
model scale. They therefore isolate the effect of replacing raw
per-step action generation with compact B-spline coefficient prediction.

\subsection{CALVIN Benchmark}
\label{sec:calvin-experiments}

CALVIN evaluates long-horizon language-conditioned behavior through sequences of five consecutive subtasks \cite{calvin2021}. We follow the standard ABC$\rightarrow$D protocol used in prior evaluations \cite{vpp2024,dawn2026}. We evaluate SMILE with VPP and DAWN to test its applicability to predictive-visual and explicit pixel-motion-conditioned denoising action experts.

\begin{table*}[t]
\centering
\caption{CALVIN ABC$\rightarrow$D long-horizon evaluation. Columns
1--5 report success rates on consecutive subtasks, and Avg. Len. denotes the mean number completed. SMILE variants use longer execution horizons while improving average sequence length. Inf./action denotes amortized inference latency per executed action.}
\label{tab:calvin-results}
\scriptsize
\setlength{\tabcolsep}{3pt}
\resizebox{\textwidth}{!}{%
\begin{tabular}{lccccccccc}
\toprule
Model & Exe. Horizon $\uparrow$ & 1 $\uparrow$ & 2 $\uparrow$ & 3 $\uparrow$ & 4 $\uparrow$ & 5 $\uparrow$ & Avg. Len. $\uparrow$ & Inf./action (ms) $\downarrow$ & Model Size $\downarrow$ \\
\midrule
VidMan \cite{vidman2024} & 1 & 0.915 & 0.764 & 0.682 & 0.592 & 0.467 & 3.42 & 295.0 & 0.7B \\
Seer \cite{seer2024} & 1 & 0.944 & 0.872 & 0.799 & 0.722 & 0.643 & 3.98 & 88.5 & 0.3B \\
Seer-Large \cite{seer2024} & 1 & 0.963 & 0.916 & 0.861 & 0.803 & 0.740 & 4.28 & 150.6 & 0.6B \\
DreamVLA \cite{dreamvla2025} & 1 & 0.982 & 0.946 & 0.895 & 0.834 & 0.781 & 4.44 & 193.2 & 0.8B \\
\midrule
DAWN \cite{dawn2026} & 10 & 0.978 & 0.916 & 0.813 & 0.752 & 0.641 & 4.10 & 32.0 & 1.2B \\
\rowcolor{SMILERow}
SMILE-DAWN (Ours) & 15 & 0.972 & 0.904 & 0.827 & 0.769 & 0.711 & 4.18 & 22.7 & 1.2B \\
\midrule
VPP \cite{vpp2024} & 10 & 0.965 & 0.909 & 0.866 & 0.820 & 0.769 & 4.33 & 19.1 & 1.8B \\
\rowcolor{SMILERow}
SMILE-VPP (Ours) & 15 & 0.962 & 0.925 & 0.885 & 0.846 & 0.798 & 4.42 & 13.0 & 1.8B \\
\bottomrule
\end{tabular}%
}
\end{table*}

\Cref{tab:calvin-results} extends the evaluation to predictive-visual and explicit pixel-motion-conditioned action experts. Both SMILE variants increase the execution horizon from 10 to 15 while improving average completed sequence length. SMILE-DAWN reduces amortized latency from 32.0 to 22.7~ms per action and increases average sequence length from 4.10 to 4.18, while SMILE-VPP reduces latency from 19.1 to 13.0~ms and increases average sequence length from 4.33 to 4.42. Most gains occur in later subtasks, where errors are more likely to accumulate across the sequence. Because the corresponding model-size scales remain unchanged, these results demonstrate that SMILE generalizes beyond VLM-centered action experts.

SMILE-VPP reaches an average sequence length of 4.42, only 0.02 below DreamVLA's 4.44, while reducing amortized inference latency from 193.2 to 13.0~ms per action relative to DreamVLA, a 14.9$\times$ speedup. This comparison places SMILE-VPP at a strong
accuracy--efficiency operating point on CALVIN.

\subsection{Real-world Experiment}
\label{sec:realworld-experiment}
\paragraph{Implementation}
We use a single-arm setup with a 7-DoF xArm7 and two RGB cameras: a fixed third-person camera and a gripper-mounted camera. We compare Evo1 and VPP with their corresponding SMILE variants. Each matched pair is initialized from the same best-performing official pretrained checkpoint and fine-tuned on our collected dataset for the same number of optimization steps, providing a matched fine-tuning budget. The dataset contains 1,000 episodes across 12 tasks.

\begin{figure}[!tbp]
    \centering
    \begin{minipage}[t]{0.40\linewidth}
        \centering
        \includegraphics[width=\linewidth,height=0.20\textheight,keepaspectratio,trim=0 0 0 490,clip]{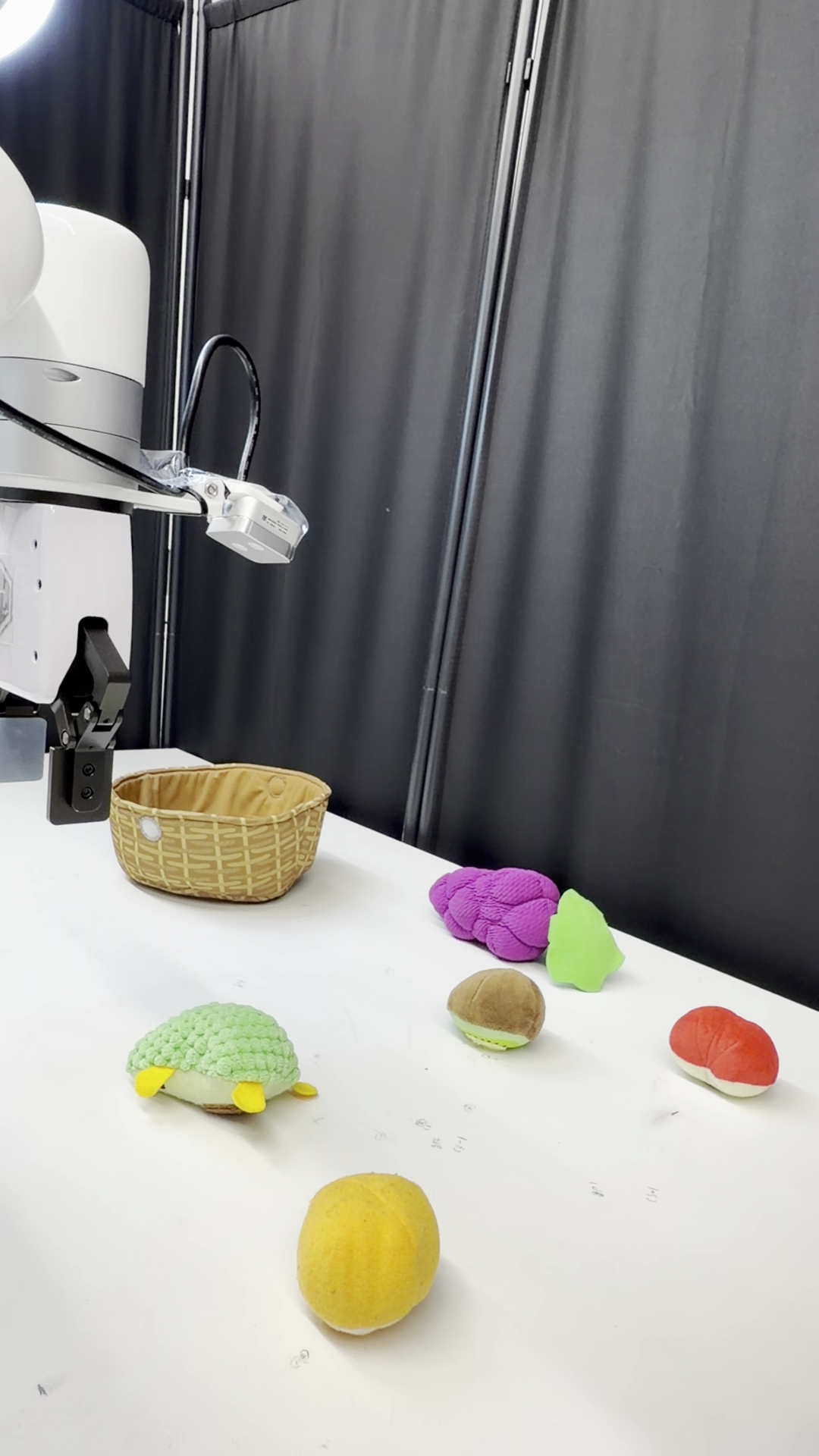}\\[-0.25em]
        {\footnotesize (a) Plain}
    \end{minipage}
    \hspace{0.05\linewidth}
    \begin{minipage}[t]{0.40\linewidth}
        \centering
        \includegraphics[width=\linewidth,height=0.20\textheight,keepaspectratio,trim=0 0 0 490,clip]{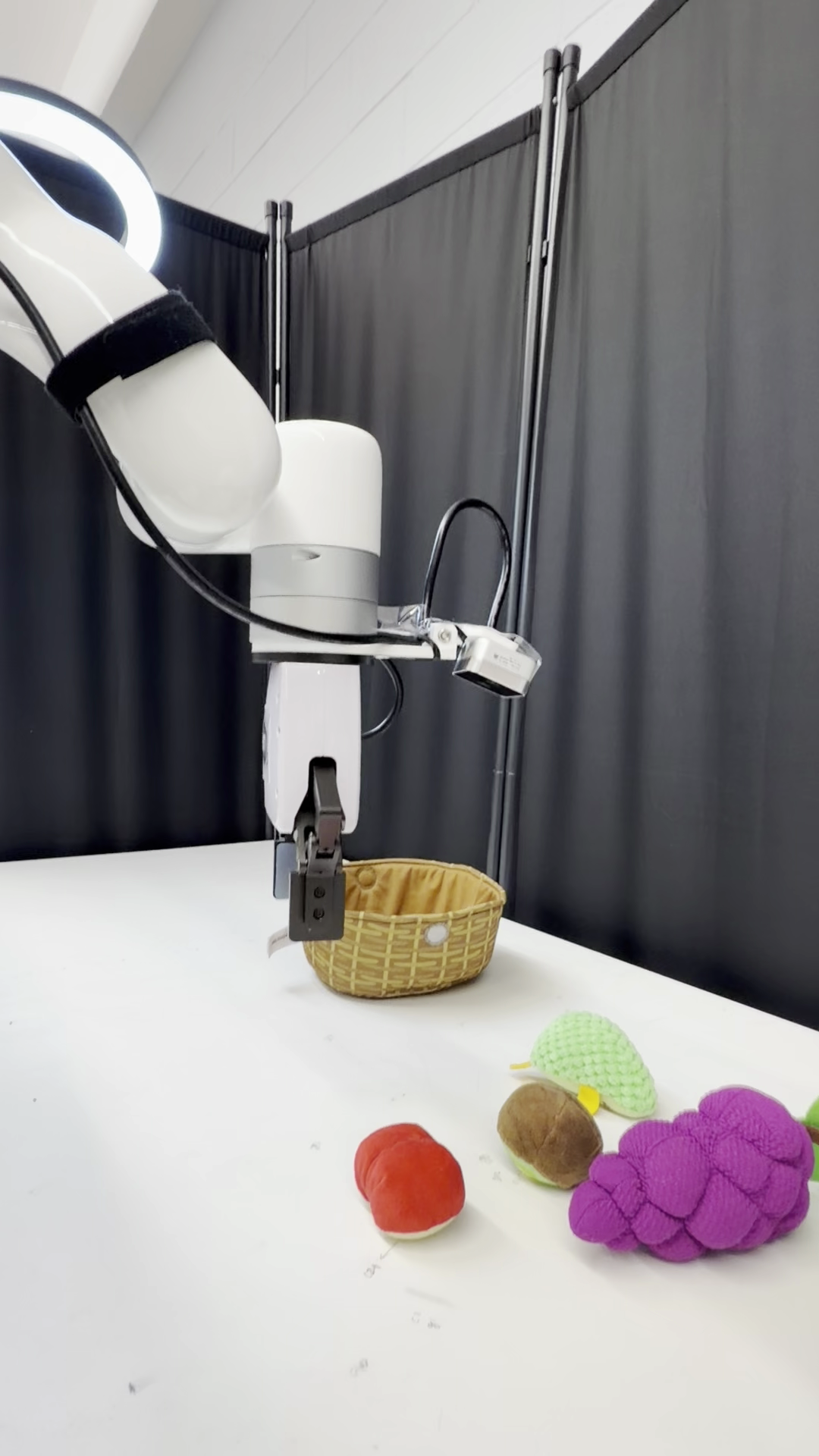}\\[-0.25em]
        {\footnotesize (b) Cluttered}
    \end{minipage}
    \caption{Representative xArm kiwi-task layouts. The plain setting uses four separated distractors, whereas the cluttered setting places three distractors tightly around the kiwi.}
    \label{fig:realworld-kiwi-layout}
\end{figure}

\paragraph{Evaluation}
The robot receives a language instruction to lift a specified object and place it into a basket. We evaluate six object categories under two arrangements, yielding 12 object-setting tasks. \Cref{fig:realworld-kiwi-layout} shows representative configurations for the prompt ``lift the kiwi and place it in the basket.'' In the \emph{plain} setting, four distractors surround the target with sufficient free space. In the \emph{cluttered} setting, three nearby distractors tightly surround the target, thereby testing accurate approach, grasping, and transport without contacting neighboring objects.

Each task is evaluated over 20 episodes with randomized initial object configurations. We use $H=6$ for Evo1 in the real-world setup rather than the LIBERO setting of $H=14$, because $H=6$ was the longest tested horizon that produced sufficiently reliable xArm rollouts.

\begin{figure}[!htbp]
    \centering
    \begin{minipage}[t]{0.31\linewidth}
        \centering
        \includegraphics[width=\linewidth,height=0.285\textheight,keepaspectratio,trim=0 0 0 200,clip]{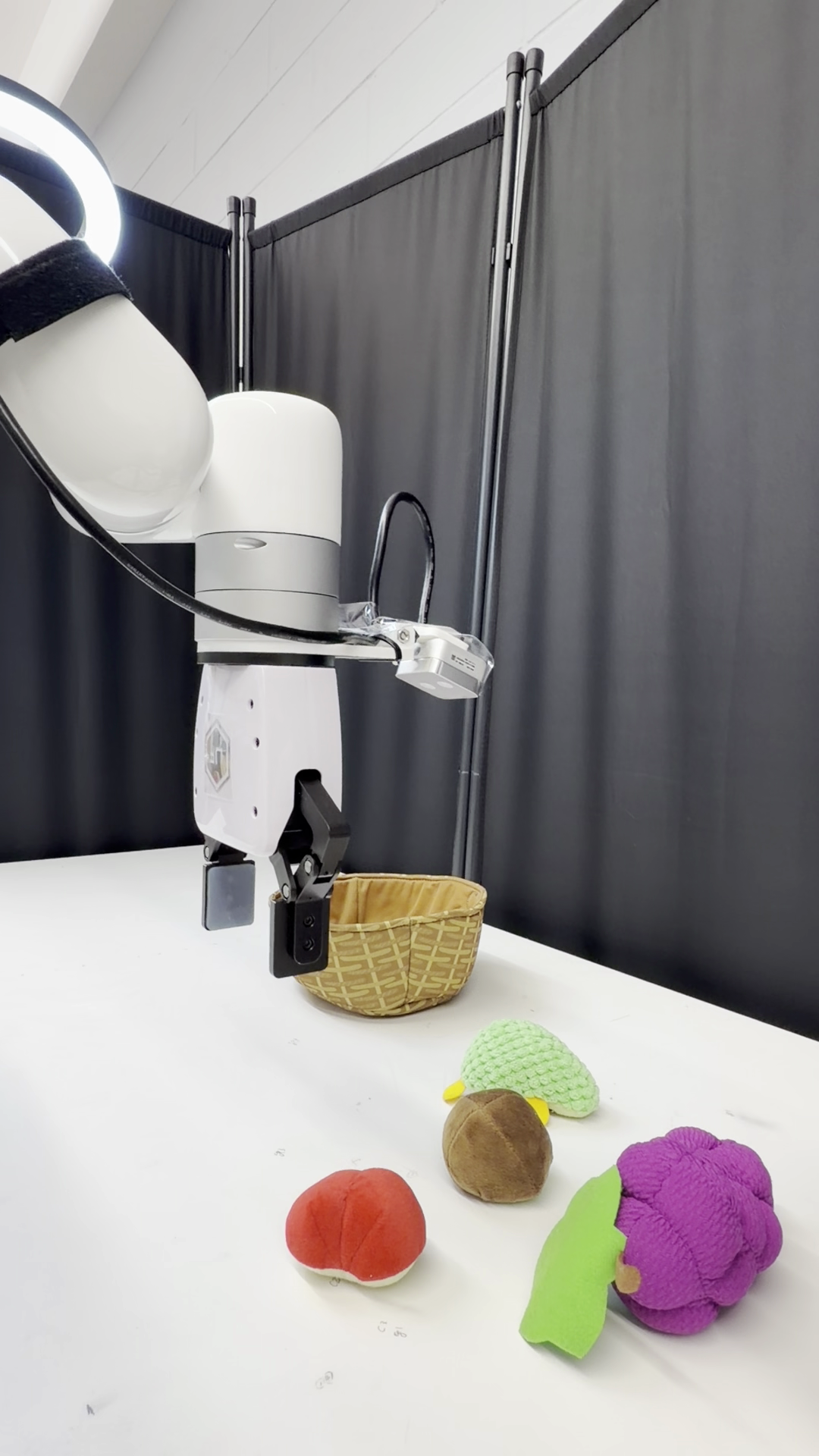}\\[-0.25em]
        {\footnotesize (a) Approach}
    \end{minipage}
    \hfill
    \begin{minipage}[t]{0.31\linewidth}
        \centering
        \includegraphics[width=\linewidth,height=0.285\textheight,keepaspectratio,trim=0 0 0 200,clip]{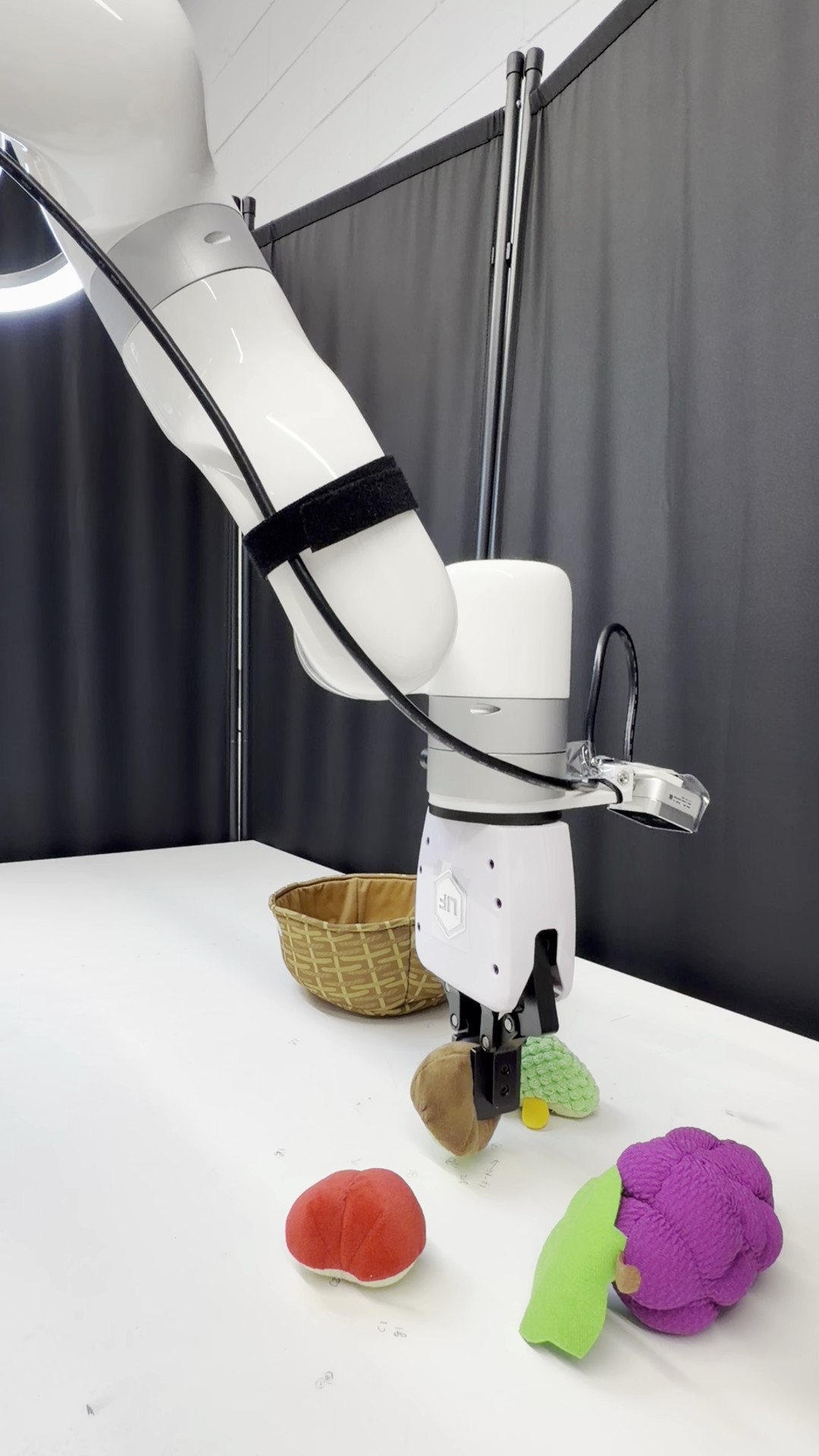}\\[-0.25em]
        {\footnotesize (b) Pickup}
    \end{minipage}
    \hfill
    \begin{minipage}[t]{0.31\linewidth}
        \centering
        \includegraphics[width=\linewidth,height=0.285\textheight,keepaspectratio,trim=0 0 0 200,clip]{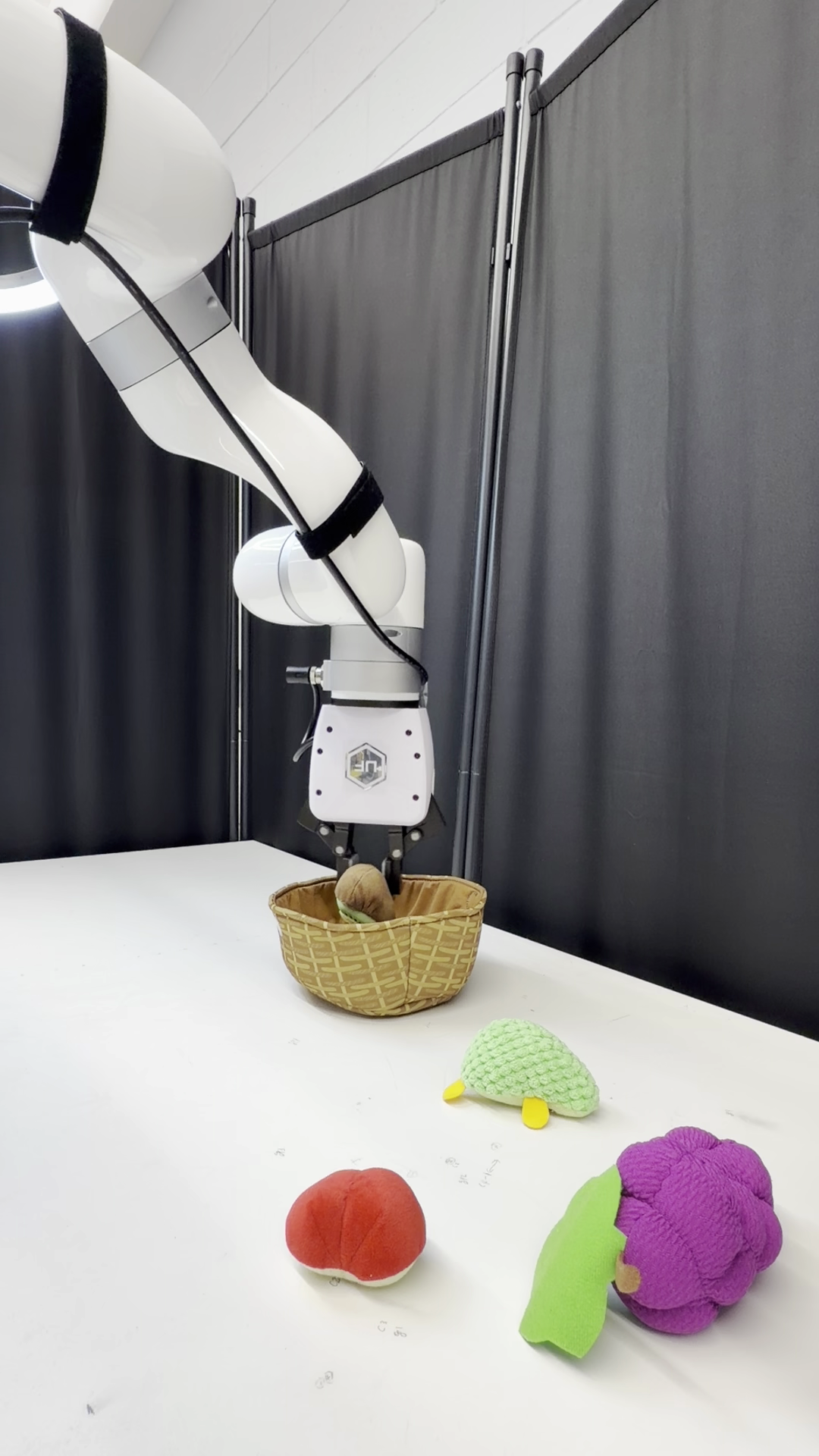}\\[-0.25em]
        {\footnotesize (c) Place}
    \end{minipage}
    \caption{SMILE-VPP cluttered-kiwi success: approach, pickup, and placement.}
    \label{fig:smile-vpp-kiwi-success}
\end{figure}

\paragraph{Results}
SMILE improves success for every object category in both settings. In the plain setting, average success increases from 26.7\% to 78.3\% for Evo1 and from 82.5\% to 98.3\% for VPP. In clutter, it increases from 8.3\% to 56.7\% and from 70.8\% to 88.3\%, respectively. \Cref{fig:smile-vpp-kiwi-success} shows a representative SMILE-VPP cluttered-kiwi success, progressing from approach to pickup and basket placement. The VPP variants achieve higher absolute success than their Evo1 counterparts, plausibly reflecting VPP's larger model scale and broader real-world pretraining. We therefore emphasize each matched baseline-to-SMILE comparison.

\begin{table*}[!t]
\centering
\caption{Real-world xArm lift-and-place results. Succ. and W.Obj. report lift$\to$placement counts for the correct and wrong objects,
respectively. Hit denotes distractor-contact episodes, and F. denotes
failures to pick up the target within three attempts, workspace exits, or freezes. Aggregate columns report execution horizon (E.H.), amortized inference latency per action (I/a), success rate (SR), wrong-object pickup rate (WPR), drop rate after any pickup (DR), and hit rate (HR) over all six objects.}
\label{tab:realworld-xarm}
\scriptsize
\setlength{\tabcolsep}{0.9pt}

\resizebox{\textwidth}{!}{%
\begin{tabular}{@{}ll*{14}{c}@{}}
\toprule
Setting & Model &
\multicolumn{6}{c}{All objects} &
\multicolumn{4}{c}{Apple} &
\multicolumn{4}{c}{Avocado} \\
\cmidrule(lr){3-8}
\cmidrule(lr){9-12}
\cmidrule(lr){13-16}
& & E.H.$\uparrow$ & I/a$\downarrow$ & SR$\uparrow$ &
WPR$\downarrow$ & DR$\downarrow$ & HR$\downarrow$ &
Succ.$\uparrow$ & W.Obj.$\downarrow$ & Hit$\downarrow$ & F.$\downarrow$ &
Succ.$\uparrow$ & W.Obj.$\downarrow$ & Hit$\downarrow$ & F.$\downarrow$ \\
\midrule

\multirow{4}{*}{Plain}
& Evo1
& 6 & 21.3 & 26.7\% & 30.0\% & 10.0\% & 27.5\%
& $6{\to}5$ & $7{\to}5$ & 5 & 2
& $9{\to}8$ & $4{\to}4$ & 4 & 2 \\

& \cellcolor{SMILERow}SMILE-Evo1
& \cellcolor{SMILERow}16
& \cellcolor{SMILERow}8.0
& \cellcolor{SMILERow}78.3\%
& \cellcolor{SMILERow}16.7\%
& \cellcolor{SMILERow}0.0\%
& \cellcolor{SMILERow}4.2\%
& \cellcolor{SMILERow}$16{\to}16$
& \cellcolor{SMILERow}$4{\to}4$
& \cellcolor{SMILERow}0
& \cellcolor{SMILERow}0
& \cellcolor{SMILERow}$19{\to}19$
& \cellcolor{SMILERow}$1{\to}1$
& \cellcolor{SMILERow}0
& \cellcolor{SMILERow}0 \\

\cmidrule(lr){2-16}

& VPP
& 10 & 19.1 & 82.5\% & 0.8\% & 5.8\% & 6.7\%
& $18{\to}16$ & 0 & 0 & 2
& $18{\to}18$ & $1{\to}1$ & 1 & 0 \\

& \cellcolor{SMILERow}SMILE-VPP
& \cellcolor{SMILERow}15
& \cellcolor{SMILERow}13.0
& \cellcolor{SMILERow}98.3\%
& \cellcolor{SMILERow}0.0\%
& \cellcolor{SMILERow}0.0\%
& \cellcolor{SMILERow}0.0\%
& \cellcolor{SMILERow}$20{\to}20$
& \cellcolor{SMILERow}0
& \cellcolor{SMILERow}0
& \cellcolor{SMILERow}0
& \cellcolor{SMILERow}$20{\to}20$
& \cellcolor{SMILERow}0
& \cellcolor{SMILERow}0
& \cellcolor{SMILERow}0 \\

\specialrule{\lightrulewidth}{0.35em}{0pt}
\specialrule{\lightrulewidth}{1.0pt}{0.35em}

\multirow{4}{*}{Cluttered}
& Evo1
& 6 & 21.3 & 8.3\% & 44.2\% & 8.3\% & 45.0\%
& $2{\to}2$ & $11{\to}9$ & 7 & 0
& $2{\to}2$ & $10{\to}8$ & 6 & 2 \\

& \cellcolor{SMILERow}SMILE-Evo1
& \cellcolor{SMILERow}16
& \cellcolor{SMILERow}8.0
& \cellcolor{SMILERow}56.7\%
& \cellcolor{SMILERow}20.8\%
& \cellcolor{SMILERow}0.0\%
& \cellcolor{SMILERow}22.5\%
& \cellcolor{SMILERow}$8{\to}8$
& \cellcolor{SMILERow}$10{\to}10$
& \cellcolor{SMILERow}2
& \cellcolor{SMILERow}0
& \cellcolor{SMILERow}$18{\to}18$
& \cellcolor{SMILERow}0
& \cellcolor{SMILERow}2
& \cellcolor{SMILERow}0 \\

\cmidrule(lr){2-16}

& VPP
& 10 & 19.1 & 70.8\% & 0.8\% & 3.3\% & 25.0\%
& $12{\to}11$ & 0 & 8 & 0
& $17{\to}17$ & 0 & 3 & 0 \\

& \cellcolor{SMILERow}SMILE-VPP
& \cellcolor{SMILERow}15
& \cellcolor{SMILERow}13.0
& \cellcolor{SMILERow}88.3\%
& \cellcolor{SMILERow}0.0\%
& \cellcolor{SMILERow}0.0\%
& \cellcolor{SMILERow}11.7\%
& \cellcolor{SMILERow}$15{\to}15$
& \cellcolor{SMILERow}0
& \cellcolor{SMILERow}5
& \cellcolor{SMILERow}0
& \cellcolor{SMILERow}$19{\to}19$
& \cellcolor{SMILERow}0
& \cellcolor{SMILERow}1
& \cellcolor{SMILERow}0 \\
\bottomrule
\end{tabular}%
}

\vspace{0.6em}

\resizebox{\textwidth}{!}{%
\begin{tabular}{@{}l*{16}{c}@{}}
\toprule
Model &
\multicolumn{4}{c}{Banana} &
\multicolumn{4}{c}{Grape} &
\multicolumn{4}{c}{Kiwi} &
\multicolumn{4}{c}{Orange} \\
\cmidrule(lr){2-5}
\cmidrule(lr){6-9}
\cmidrule(lr){10-13}
\cmidrule(lr){14-17}
& Succ.$\uparrow$ & W.Obj.$\downarrow$ & Hit$\downarrow$ & F.$\downarrow$
& Succ.$\uparrow$ & W.Obj.$\downarrow$ & Hit$\downarrow$ & F.$\downarrow$
& Succ.$\uparrow$ & W.Obj.$\downarrow$ & Hit$\downarrow$ & F.$\downarrow$
& Succ.$\uparrow$ & W.Obj.$\downarrow$ & Hit$\downarrow$ & F.$\downarrow$ \\
\midrule

Evo1
& $8{\to}6$ & $4{\to}3$ & 6 & 2
& $7{\to}6$ & $4{\to}4$ & 8 & 1
& $4{\to}4$ & $8{\to}6$ & 6 & 2
& $4{\to}3$ & $9{\to}8$ & 4 & 3 \\

\cellcolor{SMILERow}SMILE-Evo1
& \cellcolor{SMILERow}$18{\to}18$
& \cellcolor{SMILERow}$1{\to}1$
& \cellcolor{SMILERow}0
& \cellcolor{SMILERow}1
& \cellcolor{SMILERow}$16{\to}16$
& \cellcolor{SMILERow}$1{\to}1$
& \cellcolor{SMILERow}3
& \cellcolor{SMILERow}0
& \cellcolor{SMILERow}$13{\to}13$
& \cellcolor{SMILERow}$5{\to}5$
& \cellcolor{SMILERow}2
& \cellcolor{SMILERow}0
& \cellcolor{SMILERow}$12{\to}12$
& \cellcolor{SMILERow}$8{\to}8$
& \cellcolor{SMILERow}0
& \cellcolor{SMILERow}0 \\

\cmidrule(lr){1-17}

VPP
& $14{\to}12$ & 0 & 6 & 0
& $18{\to}17$ & 0 & 1 & 0
& $18{\to}18$ & 0 & 0 & 2
& $20{\to}18$ & 0 & 0 & 0 \\

\cellcolor{SMILERow}SMILE-VPP
& \cellcolor{SMILERow}$20{\to}20$
& \cellcolor{SMILERow}0
& \cellcolor{SMILERow}0
& \cellcolor{SMILERow}0
& \cellcolor{SMILERow}$19{\to}19$
& \cellcolor{SMILERow}0
& \cellcolor{SMILERow}0
& \cellcolor{SMILERow}1
& \cellcolor{SMILERow}$20{\to}20$
& \cellcolor{SMILERow}0
& \cellcolor{SMILERow}0
& \cellcolor{SMILERow}0
& \cellcolor{SMILERow}$19{\to}19$
& \cellcolor{SMILERow}0
& \cellcolor{SMILERow}0
& \cellcolor{SMILERow}1 \\

\specialrule{\lightrulewidth}{0.35em}{0pt}
\specialrule{\lightrulewidth}{1.0pt}{0.35em}

Evo1
& $3{\to}2$ & $5{\to}4$ & 12 & 0
& $3{\to}3$ & $3{\to}2$ & 14 & 0
& $1{\to}1$ & $11{\to}10$ & 8 & 0
& 0 & $13{\to}11$ & 7 & 0 \\

\cellcolor{SMILERow}SMILE-Evo1
& \cellcolor{SMILERow}$12{\to}12$
& \cellcolor{SMILERow}$2{\to}2$
& \cellcolor{SMILERow}6
& \cellcolor{SMILERow}0
& \cellcolor{SMILERow}$13{\to}13$
& \cellcolor{SMILERow}0
& \cellcolor{SMILERow}7
& \cellcolor{SMILERow}0
& \cellcolor{SMILERow}$10{\to}10$
& \cellcolor{SMILERow}$4{\to}4$
& \cellcolor{SMILERow}6
& \cellcolor{SMILERow}0
& \cellcolor{SMILERow}$7{\to}7$
& \cellcolor{SMILERow}$9{\to}9$
& \cellcolor{SMILERow}4
& \cellcolor{SMILERow}0 \\

\cmidrule(lr){1-17}

VPP
& $13{\to}12$ & 0 & 7 & 0
& $18{\to}17$ & 0 & 2 & 0
& $16{\to}16$ & $1{\to}1$ & 3 & 0
& $13{\to}12$ & 0 & 7 & 0 \\

\cellcolor{SMILERow}SMILE-VPP
& \cellcolor{SMILERow}$18{\to}18$
& \cellcolor{SMILERow}0
& \cellcolor{SMILERow}2
& \cellcolor{SMILERow}0
& \cellcolor{SMILERow}$19{\to}19$
& \cellcolor{SMILERow}0
& \cellcolor{SMILERow}1
& \cellcolor{SMILERow}0
& \cellcolor{SMILERow}$18{\to}18$
& \cellcolor{SMILERow}0
& \cellcolor{SMILERow}2
& \cellcolor{SMILERow}0
& \cellcolor{SMILERow}$17{\to}17$
& \cellcolor{SMILERow}0
& \cellcolor{SMILERow}3
& \cellcolor{SMILERow}0 \\
\bottomrule
\end{tabular}%
}
\end{table*}

In clutter, SMILE-Evo1 and SMILE-VPP reduce hit rate by 22.5 and 13.3 percentage points (pp), respectively, and reduce wrong-object
pickup rate by 23.4 and 0.8~pp. These higher baseline error rates can both reflect and compound distribution shift: jittery actions can move the robot and scene away from states represented in the demonstrations, leading to contact with distractors or wrong-object grasps. These failures can then further alter the scene and degrade subsequent action predictions.

\begin{figure}[!htbp]
    \centering
    \begin{minipage}[t]{0.48\linewidth}
        \centering
        \includegraphics[width=\linewidth,height=0.145\textheight,keepaspectratio,trim=0 200 0 560,clip]{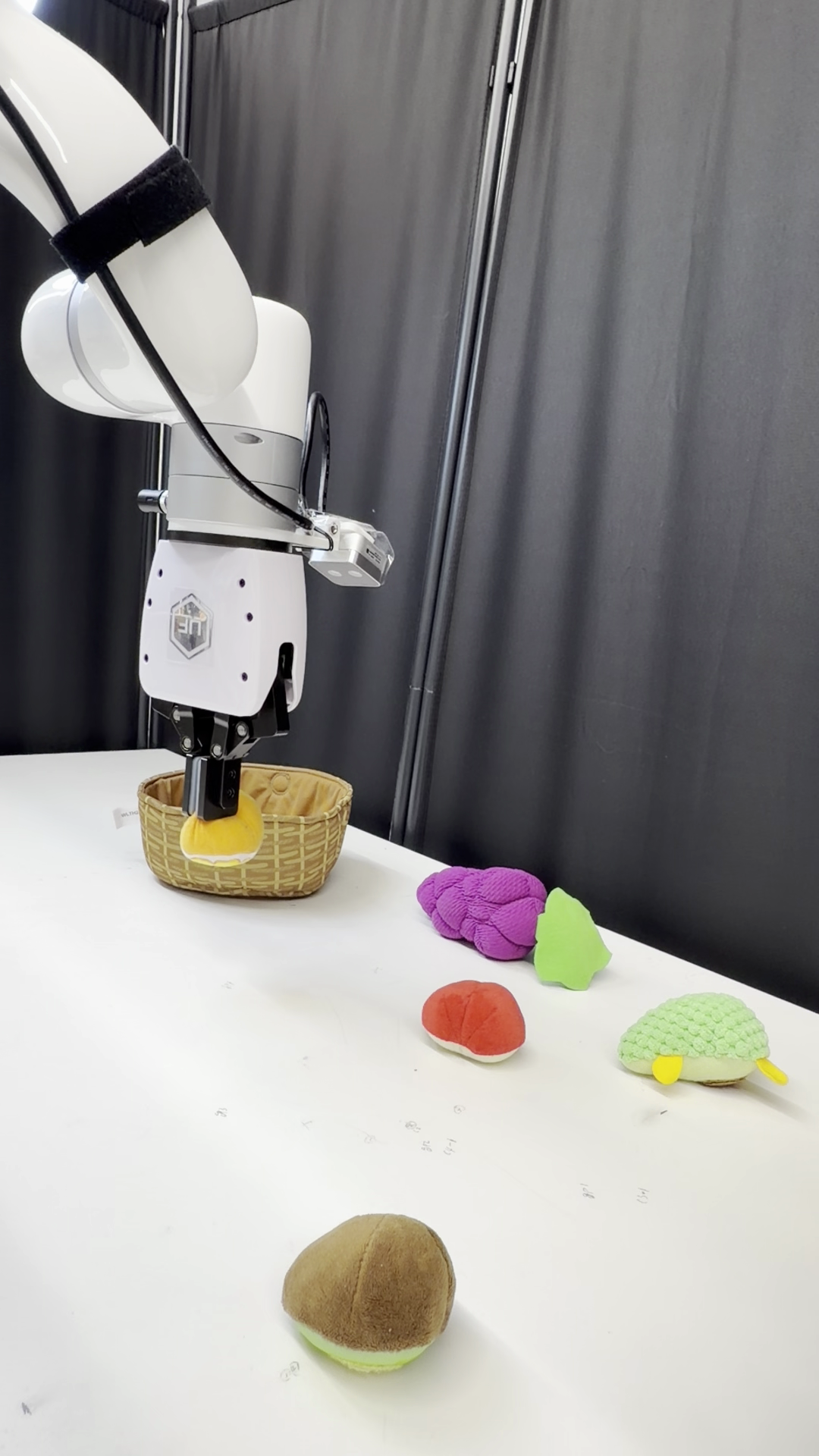}\\[-0.25em]
        {\footnotesize (a) Before drop}
    \end{minipage}
    \hfill
    \begin{minipage}[t]{0.48\linewidth}
        \centering
        \includegraphics[width=\linewidth,height=0.145\textheight,keepaspectratio,trim=0 200 0 560,clip]{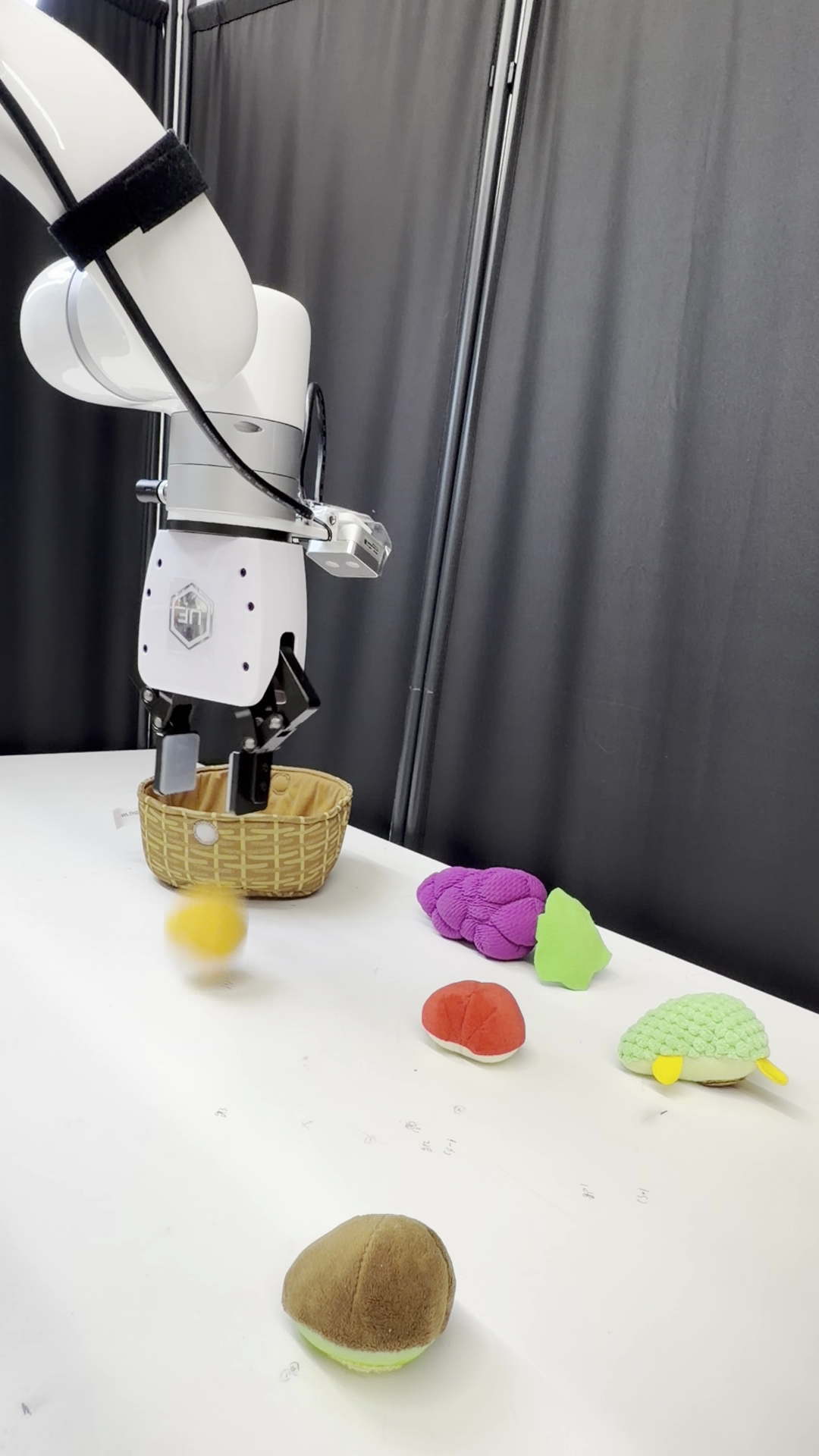}\\[-0.25em]
        {\footnotesize (b) After drop}
    \end{minipage}
    \caption{VPP orange-drop failure: the orange is held near the basket in (a) but released before reaching the basket center in (b).}
    \label{fig:vpp-orange-drop}
\end{figure}

Qualitative rollouts also show more stable gripper behavior. The baseline policies sometimes oscillate between opening and closing commands and release an object during transport; \Cref{fig:vpp-orange-drop} shows a VPP example in which the orange drops before reaching the basket center. Both SMILE variants eliminate the aggregate drops observed for the baselines, reducing drop rate in the plain setting from 10.0\% and 5.8\% to 0.0\% for Evo1 and VPP, respectively.

SMILE increases the executed horizon from 6 to 16 for Evo1 and from 10 to 15 for VPP, yielding 2.7$\times$ ($21.3{\to}8.0~ms$) and 1.5$\times$ ($19.1{\to}13.0~ms$) speedups, respectively. These results verify that smoother actions improve both task accuracy and inference efficiency in simulation and real-world environments.

\subsection{Action Smoothness Analysis}
\label{sec:action-smoothness}

To examine the mechanism behind SMILE's accuracy--efficiency gains, we compare SmolVLA and SMILE-SmolVLA on LIBERO at the same execution horizon, $H=10$, thereby matching the model-refresh frequency and isolating the effect of the action representation. All metrics are computed from normalized translation and rotation actions. For an executed sequence $a_{1:T}$, we define finite-difference acceleration as $\Delta^2 a_t=a_{t+1}-2a_t+a_{t-1}$. We average the per-episode 95th percentile of $\|\Delta^2 a_t\|$ across episodes, reporting global acceleration, non-boundary acceleration within chunks, and boundary acceleration near model-refresh transitions. We additionally report the velocity sign-change rate (Vel. ZCR), defined as the fraction of adjacent velocity steps that reverse sign, as a measure of oscillatory motion.

\begin{center}
\captionsetup{type=figure}
    \centering
    \includegraphics[width=\columnwidth]
    {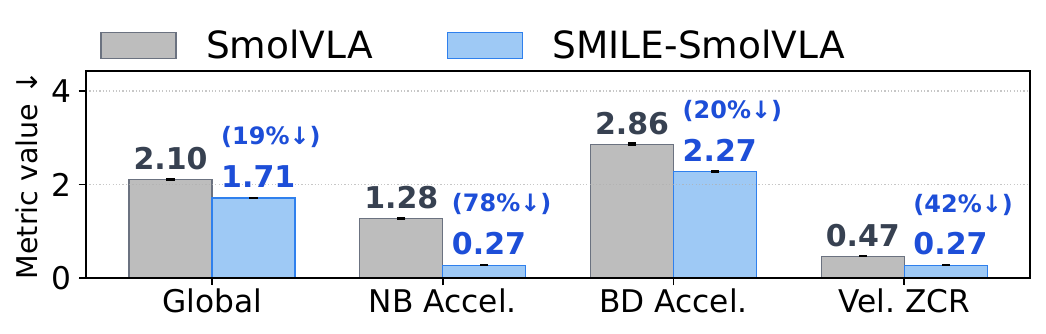}
    \caption{Aggregate LIBERO smoothness metrics. Lower values indicate smoother execution.}
    \label{fig:trrot-smoothness}
\end{center}

As shown in \cref{fig:trrot-smoothness}, SMILE-SmolVLA reduces non-boundary acceleration by 78.6\% and Vel. ZCR by 42.3\%, while reducing global and boundary acceleration by 18.6\% and 20.5\%, respectively. The much larger reduction within chunks indicates that coefficient-space generation primarily suppresses intra-chunk jitter, rather than only smoothing transitions between model calls.

\begin{figure}[!tbp]
    \centering
    \begin{subfigure}[t]{\columnwidth}
        \centering
        \includegraphics[width=\linewidth]{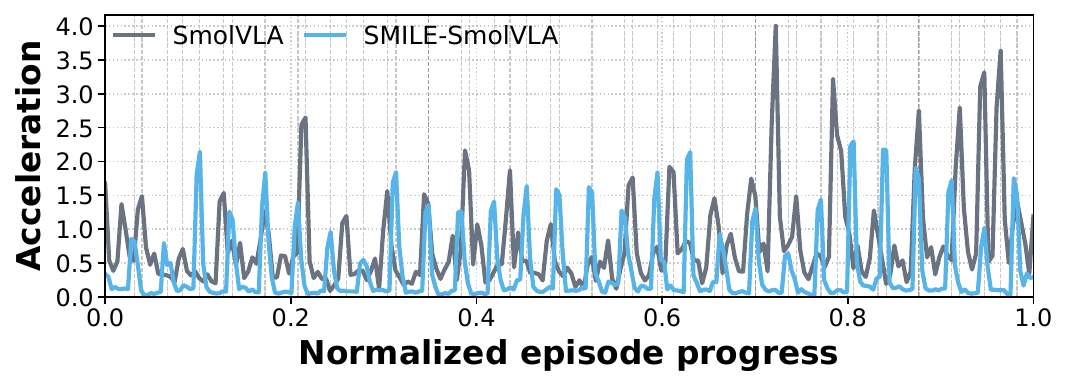}
        \caption{Task 3: put the black bowl in the bottom drawer of the cabinet and close it.}
        \label{fig:accel-task3}
    \end{subfigure}

    \vspace{0.20em}
    \begin{subfigure}[t]{\columnwidth}
        \centering
        \includegraphics[width=\linewidth]{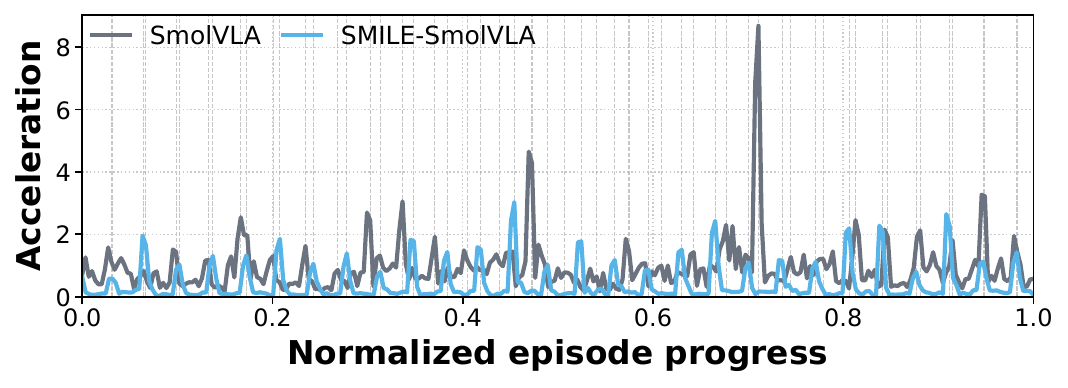}
        \caption{Task 7: put both the alphabet soup and the cream cheese box in the basket.}
        \label{fig:accel-task7}
    \end{subfigure}
    \caption{Acceleration traces for two LIBERO-10 tasks at $H=10$; dashed lines mark model-call boundaries.}
    \label{fig:accel-traces}
\end{figure}

The traces in \cref{fig:accel-traces} show frequent high-amplitude acceleration spikes for SmolVLA, both within chunks and near model-call boundaries. SMILE-SmolVLA maintains a lower acceleration baseline and suppresses high-frequency fluctuations across both task types. This qualitative evidence agrees with the aggregate metrics and supports smoother chunk execution as the mechanism behind improved long-horizon reliability.

\subsection{Ablation Study}
\label{sec:ablation-study}

This section evaluates SMILE's robustness across multiple execution horizons and identifies the optimal B-spline hyperparameters.

\begin{figure}[!t]
    \centering
    \includegraphics[width=0.85\linewidth]{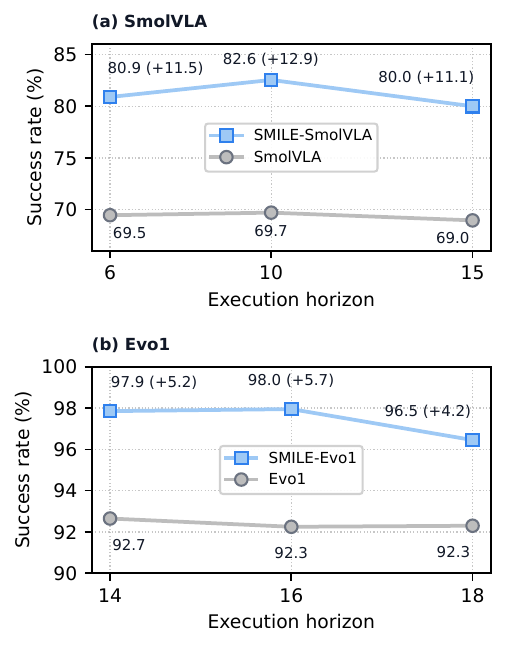}
    \caption{LIBERO success versus execution horizon. SMILE outperforms raw-action baselines across tested horizons.}
    \label{fig:horizon-robustness}
\end{figure}

\Cref{fig:horizon-robustness} shows that SMILE-SmolVLA and SMILE-Evo1 consistently outperform their corresponding raw-action baselines across multiple long execution horizons. The benefit of coefficient-space action generation is therefore robust to the horizon choice rather than arising from a single favorable operating point.

\begin{table}[!tbp]
\centering
\caption{SMILE-SmolVLA spline ablations on LIBERO. Highlighted cells mark selected hyperparameters.}
\label{tab:spline-ablation}
\scriptsize
\setlength{\tabcolsep}{2.4pt}
\renewcommand{\arraystretch}{1.05}
\begin{subtable}[t]{0.485\columnwidth}
    \centering
    \caption{Control points ($P=3$).}
    \label{tab:k-ablation}
    \resizebox{0.90\linewidth}{!}{%
    \begin{tabular}{@{}lccc@{}}
    \toprule
    $K$ & 6 & \cellcolor{SMILERow}\textbf{8} & 10 \\
    \midrule
    Succ.(\%) & 80.05 & \cellcolor{SMILERow}\textbf{82.55} & 80.75 \\
    \bottomrule
    \end{tabular}%
    }
\end{subtable}\hspace{0.01\columnwidth}%
\begin{subtable}[t]{0.485\columnwidth}
    \centering
    \caption{Spline degree ($K=8$).}
    \label{tab:p-ablation}
    \resizebox{0.90\linewidth}{!}{%
    \begin{tabular}{@{}lccc@{}}
    \toprule
    $P$ & \cellcolor{SMILERow}\textbf{3} & 5 & 7 \\
    \midrule
    Succ.(\%) & \cellcolor{SMILERow}\textbf{82.55} & 81.90 & 81.75 \\
    \bottomrule
    \end{tabular}%
    }
\end{subtable}
\end{table}

\Cref{tab:k-ablation} shows that, with cubic splines, $K=8$ control points performs best. \Cref{tab:p-ablation} shows that, with
$K=8$ fixed, the cubic degree $P=3$ outperforms the higher-degree
alternatives. A modest low-order basis is therefore sufficient;
increasing spline complexity does not improve success and may weaken
the useful smoothness bias.

\section{Conclusion}
We introduced \emph{SMILE}, an architecture-preserving B-spline action interface that replaces raw per-step action generation with compact coefficient-space prediction. By improving the temporal consistency of predicted action chunks, SMILE enables longer fixed-horizon execution with higher task accuracy and lower amortized inference latency per action. Its consistent improvements across heterogeneous VLA action experts, simulation benchmarks, and real-world manipulation demonstrate that smooth coefficient-action generation is a general mechanism for improving the accuracy--efficiency trade-off of compact VLAs, without replacing their original visual-language conditioning or denoising backbones.

\FloatBarrier

\bibliographystyle{IEEEtran}
\bibliography{smile_references_ieee}
\end{document}